\documentclass{article}

 \usepackage[dblblindworkshop, final]{neurips_2025}

\usepackage{microtype}
\usepackage{hyperref}
\usepackage{url}
\usepackage{booktabs}
\usepackage{graphicx}
\usepackage{xspace}
\usepackage{dirtytalk}
\usepackage{multirow}
\usepackage{subfigure}
\usepackage{wrapfig}
\usepackage{caption}
\usepackage{tikz}
\usepackage{tcolorbox}

\usepackage{lineno}
\usepackage{csquotes}
\definecolor{darkblue}{rgb}{0, 0, 0.5}
\hypersetup{colorlinks=true, citecolor=darkblue, linkcolor=darkblue, urlcolor=darkblue}

\newcommand{\game}{telephone game\xspace}

\newcommand{\llm}{V-LLMs\xspace}
\newcommand{\ttoi}{text-to-image models\xspace}

\newcommand{\dataset}{Telescope\xspace}
\newcommand{\gpt}{GPT-4o\xspace}
\newcommand{\newgpt}{GPT-4o-IG(20250325)\xspace}
\newcommand{\dalle}{Dall·E-3\xspace}

\title{Saying the Unsaid: Revealing the Hidden Language of Multimodal Systems Through Telephone Games}

\author{%
  Juntu Zhao \\
  Shanghai Jiao Tong University\\
  Shanghai, China\\
  \texttt{arossoneri@sjtu.edu.cn} \\
  \And
  Jialing Zhang \\
  Shanghai Jiao Tong University \\
  Shanghai, China \\
  \texttt{jialingzhang@sjtu.edu.cn} \\
  \AND
  Chongxuan Li \\
  Renmin University of China \\
  Beijing, China \\
  \texttt{chongxuanli@ruc.edu.cn} \\
  \And
  Dequan Wang\thanks{Corresponding author.} \\
  Shanghai Jiao Tong University \\
  Shanghai, China \\
  \texttt{dequanwang@sjtu.edu.cn} \\
}

\begin{document}

\maketitle

\maketitle

\begin{abstract}
Recent closed-source multimodal systems have made great advances, but their hidden language for understanding the world remains opaque because of their black-box architectures. In this paper, we use the systems' preference bias to study their hidden language: During the process of compressing the input images (typically containing multiple concepts) into texts and then reconstructing them into images, the systems' inherent preference bias introduces specific shifts in the outputs, disrupting the original input concept co-occurrence. We employ the multi-round "telephone game" to strategically leverage this bias. By observing the co-occurrence frequencies of concepts in telephone games, we quantitatively investigate the concept connection strength in the understanding of multimodal systems, i.e., "hidden language." We also contribute Telescope, a dataset of 10,000+ concept pairs, as the database of our telephone game framework. Our telephone game is test-time scalable: By iteratively running telephone games, we can construct a global map of concept connections in multimodal systems' understanding. Here we can identify preference bias inherited from training, assess generalization capability advancement, and discover more stable pathways for fragile concept connections. Furthermore, we use Reasoning-LLMs to uncover unexpected concept relationships that transcend textual and visual similarities, inferring how multimodal systems understand and simulate the world. This study offers a new perspective on the hidden language of multimodal systems and lays the foundation for future research on the interpretability and controllability of multimodal systems.


\end{abstract}

\section{Introduction} \label{sec:intro}
Recent multimodal systems, particularly closed-source ones~\citep{hurst2024gpt, step1v, bai2025qwen2}, have made significant advances, e.g., the newest \gpt with Image Generation~\citep{gpt4oimg} (abbreviated as \newgpt).
However, because of these systems' closed features, closed data, and even closed architectures, we are unable to study the systems' understanding of the world using methods based on training.
Therefore, test-time methods are urgently needed.


The hidden language reflects the connection strength between concepts within multimodal systems~\citep{chefer2023hidden}, offering insight into how they understand the world.
While prior training-based methods explored it via internal features~\citep{chen2023stair, chefer2023hidden, ghandeharioun2024patchscopes}, the rise of closed-source models renders such access impossible.
Hence, we investigate the hidden language of multimodal systems at test time.

\begin{figure}[t]
    \centering
    \includegraphics[width=0.9\textwidth]{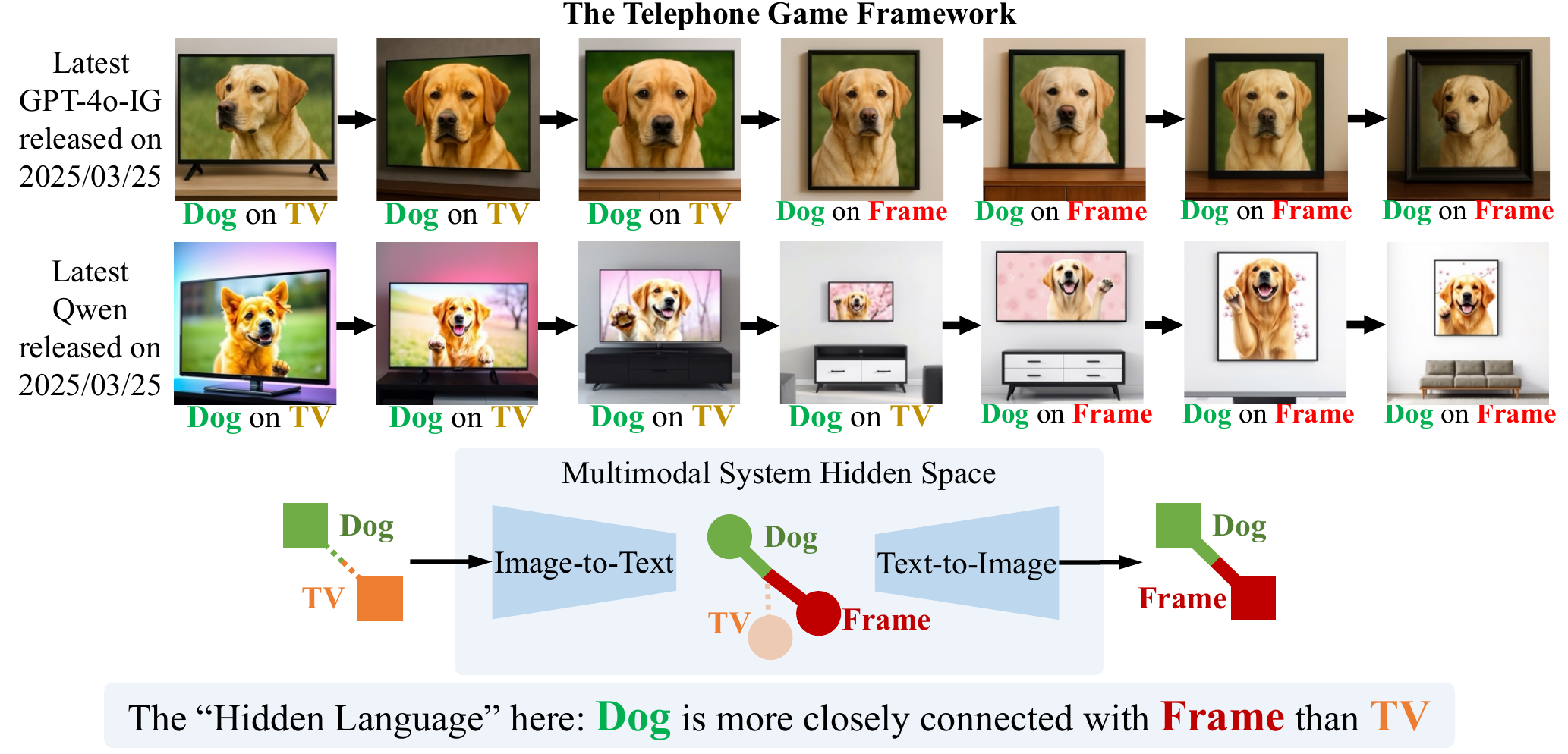}
    \caption{Example 5-round telephone games using the latest SOTA multimodal systems released on 2025.3.25. In each image reconstruction, the system prefers stronger nearby concept connections in multimodal systems' understanding, then changing the outputs. (Extended results on this example are provided in Appendix~\ref{sec:appendix/visual}), and more examples can be found in Appendix~\ref{sec:exp/gen4o}}
    \label{fig:teaser}
\end{figure}

We innovatively propose to strategically leverage the multimodal systems' preference bias to study their hidden language at test time.
Multimodal systems are trained to fit textual and visual representations of the same scenes, which typically involves multiple interrelated concepts.
Sufficient training strengthens these concept connections in systems' hidden understanding space (abbreviated as hidden space), while limited training weakens them.
Therefore, imbalanced training data brings different concept connection strengths, i.e., hidden language.
As illustrated in Figure~\ref{fig:teaser}, during image-to-text compression, the systems prefer to discard weakly connected concepts; during text-to-image reconstruction, the systems prefer strongly connected concepts~\citep{zhao2024lost}, even with the latest SOTA \newgpt.
These preference biases will lead to changes in input concepts, thereby disrupting their co-occurrence in the output scene.

\begin{figure}[t]
    \centering
    \includegraphics[width=0.9\textwidth]{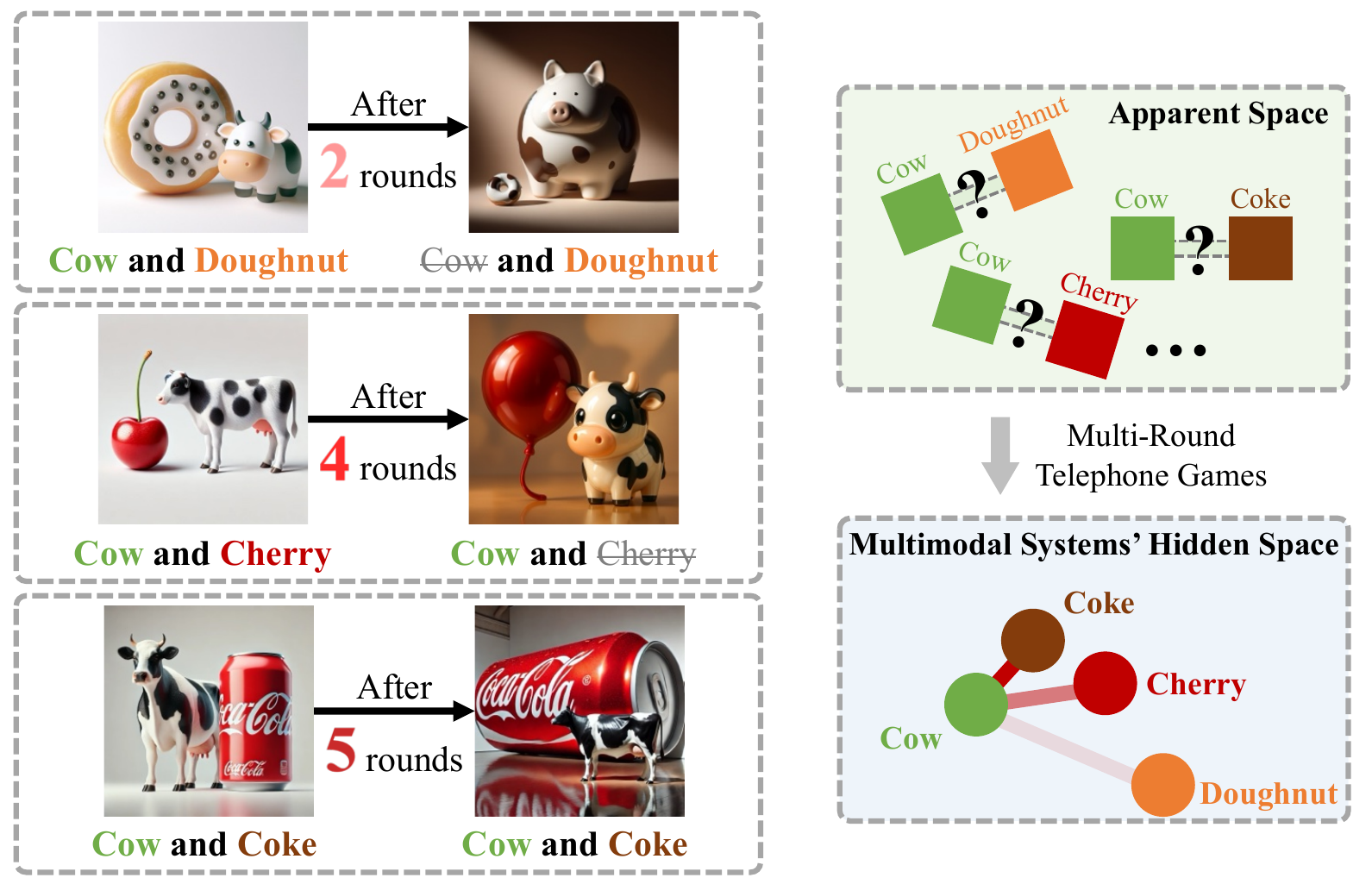}
    \caption{The longevity of concepts combinations in the \game (i.e., their co-occurrence frequency) quantitatively reflects the concept connections in multimodal systems' hidden space, termed the "hidden language." (Lighter color means the weaker connection)}
    \label{fig:connect}
\end{figure}

In this paper, we innovatively propose a \textbf{test-time} framework based on multi-round \textbf{\game} to leverage this preference bias, a plug-and-play method involving multiple cycles of image reconstruction.
As the \game progresses, fragile concept combinations gradually degrade, revealing their fragile connection strength in systems’ understanding.
And we quantify the connection strength (i.e., hidden language) using the concept co-occurrence frequency in the \game.
As shown in Figure~\ref{fig:connect}, a higher co-occurrence frequency indicates a stronger concept connection.
This metric captures both the training bias and generalization capability: Stronger generalization enables consistent responses to similar patterns, corresponding to a uniform connection strength distribution.

We also contribute \dataset, a dataset consisting of 10,000+ concept pairs derived from 150 common visual concepts, primarily covering basic spatial relations (e.g., \say{$A$ adjacent to $B$}) and some complex interactions (e.g., \say{$A$ displayed on TV screen}).
Leveraging the \game and \dataset, we propose a scalable test-time probing framework for the hidden language of multimodal systems:
Each new \game iteration tends to reveal new concept connections, and as test-time compute scales up, we progressively build a detailed \say{world map} of the multimodal hidden language.

In this way:
\textbf{(1)} We uncover key terms associated with a concept in multimodal systems' understanding, revealing the training bias (which combinations are better-trained or not) and the systems' generalization capability;
\textbf{(2)} By analyzing connection strengths across multiple pathways, we can identify intermediate concepts to enhance concept connections to promote the co-occurrence of discordant concepts;
\textbf{(3)} Reasoning-LLMs help to understand how the these connection strengths interprets physical-world laws, revealing unexpected relationships beyond textual and visual similarities.

\textbf{Here, we summarize our contributions:}
\begin{itemize}
    \item \textbf{Test-time Telephone Game Framework}: We innovatively propose to reveal the hidden language of multimodal systems using the framework of telephone game and the concept co-occurrence frequency at test time;

    \item \textbf{Telescope Dataset}: We contribute the Telescope, a database for systematic \game probing on multimodal systems' hidden language;

    \item \textbf{Test-Time Scalable Framework}: We keep on creating an increasingly comprehensive hidden language world map of multimodal systems in a scalable way.
\end{itemize}

\section{Related Works}
\paragraph{MultiModal Systems}
Recent advances in multimodal intelligence systems~\citep{lu2019vilbert, baltruvsaitis2018multimodal, xie2024large, guo2019deep, li2023blip, li2022blip, tan2019lxmert} have shown great ability in processing cross-modal information.
Modular pipeline frameworks and autoregressive systems represent two typical paradigms in multimodal architecture: the former leverages \llm~\citep{hurst2024gpt, alayrac2022flamingo, liu2023visual, wu2024next} as core components to construct complex cross-modal connections, while the latter unifies different modalities through sequential modeling within a shared hidden space~\citep{team2024chameleon, chern2024anole, li2025fractal}.
Furthermore, in recent years, the emergence of \gpt~\citep{hurst2024gpt} marks a shift toward a fully black-box system paradigm, particularly with the latest version that integrates image generation ability into \gpt~\citep{gpt4oimg}.
However, as multimodal systems' internal structures become more complex and black-box, making the way they understand the world and their preferences harder to interpret.

\paragraph{Hidden Language}
In traditional machine learning, researchers could intuitively examine a model’s hidden language using tools like attention maps~\citep{vaswani2017attention}, or Principal Component Analysis(PCA)~\citep{hotelling1933analysis}.
As deep learning and large-scale models emerge, it becomes common to train lightweight probing models on embeddings to better understand internal representations~\citep{alain2016understanding, chefer2023hidden, ghandeharioun2024patchscopes, derby2018using, chen2023stair, frank2021vision}.
However, with the emergence of many closed-source systems~\citep{gpt4oimg} today, we no longer have access even to basic token-level representations.
As a result, we are forced to infer the hidden language of multimodal systems directly from their apparent-level outputs, e.g., textual or visual outputs. This constraint effectively pivots the research paradigm from direct internal inspection to a form of behavioral analysis, where the model's observable responses to carefully crafted inputs become the primary source of evidence for its representation structures.

\section{Framework: Telephone Game } \label{sec:architecture}
\begin{figure}[t]
    \centering
    \includegraphics[width=0.9\textwidth]{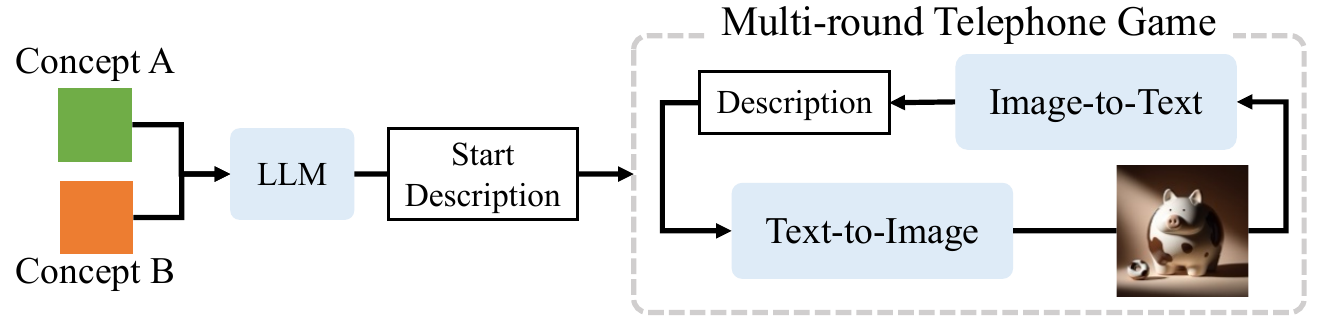}
    \caption{The workflow of \game. LLMs convert concept pairs into the start description for \game. Then it enters the cycle of text-to-image and image-to-text.}
    \label{fig:game}
\end{figure}

This section introduces our test-time \game framework for uncovering the hidden language of multimodal systems,
the concept co-occurrence frequency metric for quantifying hidden language,
and the \dataset dataset used to support systematic evaluation.

\subsection{Telephone Game} \label{sec:game}
We propose to use \game to study the concept connections in multimodal systems' hidden space, revealing the hidden language.
Our \game framework involves two key processes:
\begin{itemize}
    \item Image to Text: When compressing images into text, systems prefer to read more strongly connected concepts in their understanding over those strictly faithful to visual facts. For example in Figure~\ref{fig:connect}, it reads a cow as a pig;
    \item Text to Image: When reconstructing text to images, systems prefer to create more strongly connected concepts understood from the text to synthesize the visual output. For example, in Figure~\ref{fig:connect}, it creates a balloon instead of a cherry.
\end{itemize}

These two preferences introduce the different concept connection strengths in multimodal systems, representing the systems' hidden language.
To reveal this hidden language at test time, without accessing model parameters, we link the above processes and use changes in apparent space (e.g., image descriptions) to explore the concept connection strengths in the hidden space, as shown in Figure~\ref{fig:game}.
Changes in a single reconstruction may not be apparent in the observable space, e.g., generating a visual resembling both a cow and a pig in Figure~\ref{fig:connect}.
Moreover, concepts with fragile connections (rather than absent) may not exhibit a crash initially, but as the cycle progresses, the resulting offsets gradually become apparent.
Given these issues, we naturally use the multi-round \game to amplify the changes.

In our experiments, for fully integrated multimodal systems like the latest \gpt~\citep{gpt4oimg}, we directly utilize the system to perform both text-to-image and image-to-text processes.
For multimodal systems composed of separate components, we assemble them using \llm and \ttoi from the same institution, treating all components as a unified system.
All of our instruction prompts are available in Appendix~\ref{sec:appendix/prompt}.

\subsection{Co-occurrence Frequency} \label{sec:metric}
Modern multimodal systems rely on text and visuals, where semantic or visual similarity can seemingly reflect the systems' hidden language.
However, as shown in Figure~\ref{fig:connect}, the observed concept connection strengths, e.g., cows and coke, contradict this intuition, highlighting the need for a new metric to capture the hidden language more accurately.

The tendency of concept pairs to co-occur across multi-round \game offers key insight into the hidden language of multimodal systems.
Therefore, we propose to use the co-occurrence frequency of different concepts in the multi-round \game as a direct measure of connection strength in the system’s hidden space, reflecting the hidden language.
In a $n$-round \game, the co-occurrence frequency of the concept pairs \say{$A$ and $B$} is defined as:
\begin{equation}
    F(A, B) = \frac{\sum_{i=1}^r\sum_{j=1}^n \mathcal{I}_{i,j}(A, B)}{r \times n}
\end{equation}

where $r$ means we repeat a \game for $r$ times, and $\mathcal{I}_{i,j}(A, B)$ represents whether $A$ and $B$ co-occur in the output of the $j$-th round of the $i$-th \game judged by LLMs (the instruction prompt is available in Appendix~\ref{sec:appendix/prompt}).
In this study, we choose the image description to analyze the concept co-occurrence frequency.

Note that an implicit limitation lies in the number of rounds.
When calculating the metric correlation, we exclude pairs with a co-occurrence frequency of 1.0, as we cannot run an infinite-round \game to get a true co-occurrence \say{probability.}
And in Section~\ref{sec:exp/connection}, we demonstrate some interesting phenomena that emerge as the round number increases.

\begin{figure}[t]
    \centering
    \includegraphics[width=0.9\textwidth]{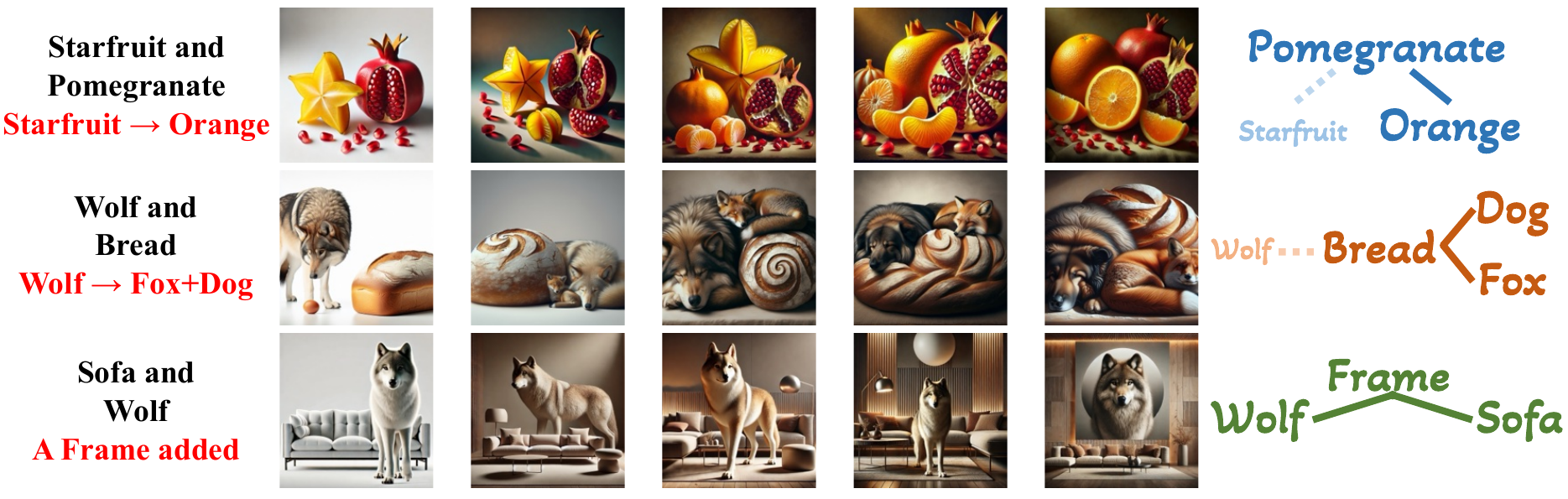}
    \caption{Visualization of several \game examples, reflecting which concepts are connected strongly or weakly in the hidden space of multimodal systems, as well as the intermediary concepts that build stronger connections. For more results, see Appendix~\ref{sec:appendix/visual}.}
    \label{fig:exp_visual}
\end{figure}

\subsection{Dataset: \dataset} \label{sec:dataset}
We also contribute a dataset, \dataset.
We collect 150 common concepts as the basic \say{vocabulary} of the hidden language.
In this study, we focus on combinations of 2 concepts, with plans to explore complex combinations of more concepts in future work.

In the main portion of the dataset, these concepts form 11,175 concept pairs, each involving arranging 2 concepts side by side, called the simple-pattern.
The other portion of the dataset represents more complex combinations of concepts, called complex-pattern:
We investigate 3 interesting visual fusion strategies: displaying Concept $A$ on a TV screen, creating Concept $A$ in the visual style of Van Gogh, and constructing Concept $A$ using wood as a material.
Unlike simple-patterns, they involve interactions between different concepts.
The \dataset dataset allows us to explore how the system establishes its hidden language in the hidden understanding space. 

\subsection{LLMs as "MLPs"}
Concept connections, viewed as the hidden language, not only reveal training data biases and generalization capabilities, but also open the door to deeper inquiry:
\textbf{\textit{What further insights into the system’s hidden logic about how it understands and simulates the real-world physical laws might emerge?}}
In this study, we abstract the text as a special \say{embedding} bridging hidden features and observed pixels.
In conventional deep learning, linear probes (e.g., MLPs) interpret embeddings to reveal internal logic;
analogously, we employ Reasoning-LLMs as cognitive probes to parse textual evolution across rounds.
These analyses uncover implicit constraints on real-world laws beyond observed-level correlations (e.g., textual or visual similarity), suggesting that multimodal systems attempt to simulate human world laws and causal relationships. The experimental details and results can be found in our Appendix~\ref{sec:llm_as_mlp}.

\newpage

\section{Experiments} \label{sec:exp}

\subsection{Model and Dataset}
\paragraph{Model}
Our primary experiments utilize OpenAI’s multimodal system, recognized as SOTA.
Here, we use the system composed of \gpt~\citep{hurst2024gpt} and \dalle~\citep{openai2023dalle3}, which is also the configuration used in OpenAI's official products.
Preliminary results show that even for simple tasks, i.e., a single concept or two identical concepts, after a 5-round \game, the original concepts exhibit crashes (disappearing or transforming, might because of the emergence of irrelevant concepts) at rates of 26.4\% and 24.4\%, respectively, highlighting the significant bias in multimodal systems, forming a key basis for our framework, as analyzed in Section~\ref{sec:architecture}.
As for the latest \newgpt, we present the experimental results using the web version of the tool in Appendix~\ref{sec:exp/gen4o}.

In Section~\ref{sec:exp/correlation}, we also analyze the hidden language of various multimodal systems derived from different sources, including:
(1) StepFun~\citep{step1v}: Step1V and Step1X, (2) Qwen~\citep{bai2025qwen2}: Qwen2.5-VL and Wanx2.1.
Simple open-source systems are excluded, as closed-source systems now far outperform open-source ones.
Open-source systems face limitations in knowledge acquisition and input text length.
For an extreme example, if the text is as short as only 3 words, e.g., \say{$A$ and $B$}, the exposure of issues would largely depend on chance.
But we can use its open-access features to validate multimodal systems' preference for strong concept connections, see Appendix~\ref{sec:appendix/opensource}.

\paragraph{Dataset}
Our dataset, \dataset, consists of over 10,000 concept pairs.
Due to time and cost constraints, we present results on a refined subset, with selection strategies detailed in the following sections.
However, we will not stop our experiments, and we are committed to continuously expanding the global map of multimodal systems' hidden language.

\subsection{Correlation} \label{sec:exp/correlation}

\begin{table}[htbp]
\centering
\begin{tabular}{@{}c|ccc@{}}
\toprule
                    & Co-occur vs Semantic & Co-occur vs Visual  & Semantic vs Visual \\
Metric Correlation $\uparrow$ & 0.046                & -0.178              & 0.041                \\ \midrule
                    & OpenAI vs StepFun     & OpenAI vs QWen & StepFun vs QWen \\
System Correlation $\uparrow$ &  0.506             & 0.475               &   0.503     \\ \bottomrule
\end{tabular}
\caption{Pearson Correlation Coefficients among the 3 metrics and different multimodal systems. Using the OpenAI system to calculate co-occurrence frequency (Co-occur) for metric comparison and our Co-occur metric for analyzing hidden language correlations across different systems, we find that semantic and visual similarities fail to capture the hidden language, while hidden languages across different systems show good correlation.}
\label{tab:metric_correlation}
\end{table}

The concepts' co-occurrence frequency reflects their connection strength in hidden space.
We are particularly interested in whether this phenomenon can be explained by existing similarity metrics, such as semantic and visual similarity. First, we detail the setup:
\begin{itemize}
    \item Metric: To explore the correlation of different metrics, we use CLIP model~\citep{radford2021learning} for semantic embeddings and ResNet-50~\citep{he2016deep} for visual embeddings, and compute concepts similarity between embeddings;

    \item Models: To explore the correlation of hidden languages in different systems, we implement 3 multimodal systems: OpenAI, StepFun, and QWen;

    \item Dataset: Given the substantial cost, we rank the simple-pattern concept pairs in \dataset by the average of their semantic and visual similarities, and uniformly sample 400 pairs.

    \item Telephone Game: For each pair, we repeat a 5-round \game for 3 times;

    \item Focus: Here, we focus on the connections between the original input concepts. The new concepts emerging during the \game will be analyzed in Section~\ref{sec:exp/connection}.
\end{itemize}

\newpage

\subsubsection{Metric Correlation}
We conduct quantitative correlation analysis to examine relationships between different metrics.
Among the 400 concept pairs, 246 experience concept crashes.
As discussed in Section~\ref{sec:metric}, due to the round number limitation, concepts with a co-occurrence frequency of 1.0 (i.e., no crash) are excluded from metric correlation analysis, as these frequencies might be unreliable.
Therefore, the correlation is computed based on the 246 crashed pairs.
We also present an intuitive visualization of their correlation in Appendix~\ref{sec:appendix/correlation}

For each of the 246 concept pairs, we compute semantic similarity, visual similarity, and co-occurrence frequency between the 2 concepts in the pair.
We calculate the Pearson Correlation Coefficients~\citep{pearson1895vii} by measuring pairwise correlations between the three 246-length lists.
As shown in Table~\ref{tab:metric_correlation}, semantic and visual similarities fail to capture concept connections, underscoring the need for metrics like our co-occurrence frequency.

\subsubsection{System Correlation}
We further examine whether the hidden languages of multimodal systems from different sources are correlated.
We run the \game on 400 concept pairs using StepFun and QWen systems, repeating each experiment 3 times.
Since the same metric is compared across systems, the round number limitation does not apply.

As reported in Table~\ref{tab:metric_correlation}, we observe a moderate correlation between the hidden languages of different multimodal systems, indicating potentially consistent concept connections in the hidden space.
This consistency significantly surpasses correlations based on semantic or pixel similarity, suggesting that our co-occurrence frequency metric captures deeper concept connections in the hidden space of multimodal systems.
Notably, this aligns with the Platonic Representation Hypothesis~\citep{huh2024position}: as multimodal systems scale, their internal representations tend to converge toward modeling the joint statistical structure of real-world events, despite differences in architecture, data, or training methods.

\subsection{Hidden Language: the Connections Between Concepts} \label{sec:exp/connection}
\begin{figure}[t]
    \centering
    \includegraphics[width=0.9\textwidth]{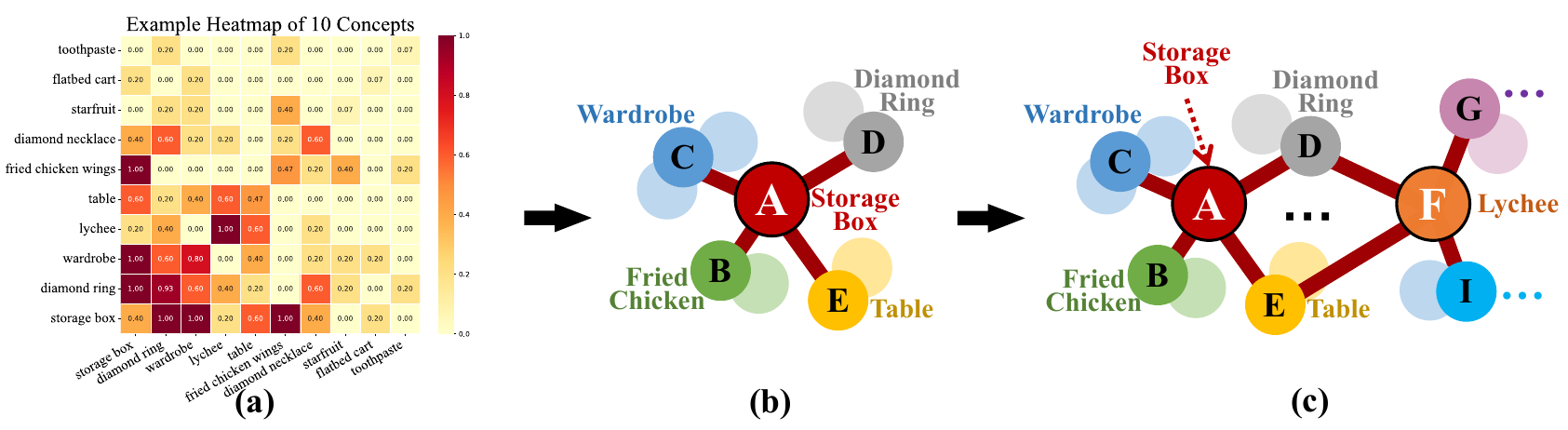}
    \caption{Our scalable framework: (a) Basic Connection: we reveal concept connections and identify nearby keywords; (b) Local Connection: repeated telephone games establish a local graph around a concept; (c) Global Connection: increasing telephone games connect local structures, forming a comprehensive "world map" of the multimodal hidden language!}
    \label{fig:world_map}
\end{figure}

Our framework has excellent dynamic scalability.
As test time compute scales, we gradually construct an increasingly comprehensive \say{world map} of the system's hidden language!

\paragraph{Basic Connection}
The map starts from our 150-concept vocabulary in \dataset.
We visualize connections for 10 example concepts in Figure~\ref{fig:world_map}(a), where the color intensity reflects connection strength, highlighting better-trained pairs, and as it scales up, revealing the system's generalization progress:
stronger generalization capability will lead to more uniform heatmap distributions.
We also present some visualization results in Figure~\ref{fig:exp_visual}.

\paragraph{Local and Global Connection}
In our framework, each new \game tends to build new connections, linking existing and newly emerging concepts.
As connections accumulate, genuine neighbors consistently reappear and occasional ones are submerged, gradually shaping a stable and accurate local structure (Figure~\ref{fig:world_map}(b)).
As the framework further expands, connections emerge between distinct local structures, gradually weaving a unified global network (Figure~\ref{fig:world_map}(c)). This integration bridges isolated local clusters and enriches the system by enabling cross-domain information exchange.
In Appendix~\ref{sec:appendix/map}, we provide a detailed explanation of the building details of such a graph structure in the \say{hidden language world map} of multimodal hidden space.

\begin{figure}[t]
    \centering
    \includegraphics[width=0.9\textwidth]{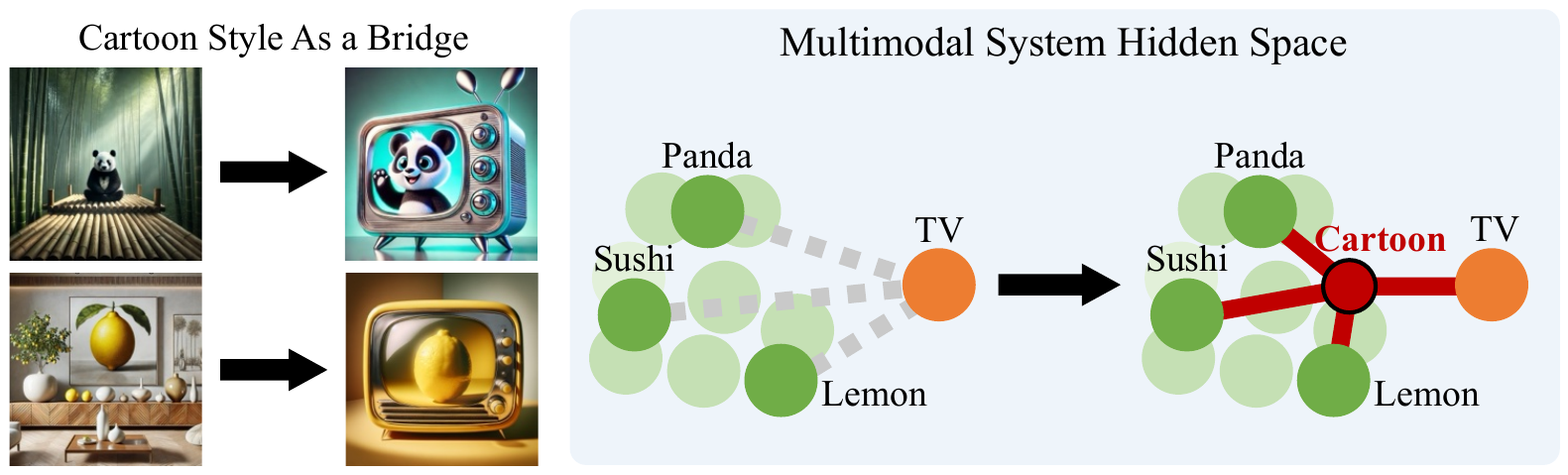}
    \caption{The intermediary node forms stable pathways between weakly connected concepts, making previously unstable combinations more reliably appear in generated images.}
    \label{fig:complex_pattern}
\end{figure}

\subsection{Complex Pattern and Concept Bridge} \label{sec:exp/complex_pattern}

\begin{minipage}[t]{0.37\textwidth}
    \vspace{0pt}
    \centering
    \vspace{0pt}
    \begin{tabular}{@{}lc@{}}
    \toprule
    Pattern              & Crash Ratio \\ \midrule
    Van Gogh Style & 0.767       \\
    Wood Texture   & 0.560        \\
    TV             & 0.740       \\
    TV (\textit{improved})  &  \textbf{0.427}    \\ \bottomrule
    \end{tabular}
    \captionof{table}{The crash ratio of our complex pattern and the "bridged improved" results on Pattern TV.}
    \label{tab:complex_pattern}
\end{minipage}
\hfill
\begin{minipage}[t]{0.6\textwidth}
As detailed in Section~\ref{sec:dataset}, our \dataset includes 450 complex pattern concept pairs, derived from 150 common concepts across 3 patterns.
We run the \game on these pairs using the OpenAI system.
In Table~\ref{tab:complex_pattern}, we report the concept crash ratios.
Among the patterns, \say{Van Gogh Style Painting} (abbreviated as \say{Van Gogh Style}) and \say{Displaying on a TV Screen} (abbreviated as \say{TV}) show notably more fragile connections than simple-pattern in Section~\ref{sec:exp/correlation}.
For example, the system has less fitting to the scenario \say{displaying concepts on a TV screen} during the training phase.
\end{minipage}

It shows that, after learning a limited number of these scenarios, the system has not developed robust generalization capabilities to extend this understanding to other concepts.

We explore bridging in the hidden space by introducing intermediary concepts, e.g., \say{Cartoon Style} or \say{Advertising Format}, in Pattern TV.
For each intermediary, we conduct experiments on 150 concepts to build a \say{TV}-centered hidden language local map like Figure~\ref{fig:world_map}(b).
This map helps identify effective intermediary nodes for stabilizing fragile connections:
using the concept\say{Cartoon Style}, we form more stable pathways, as shown in Table~\ref{tab:complex_pattern} and Figure~\ref{fig:complex_pattern}.
As the map expands, we are committed to discovering more such bridges to enhance superalignment between multimodal system inputs and outputs.






\section{Discussion}
In this study, we quantify the concept connection strengths in multimodal systems' hidden space, also termed \say{hidden language}, using our \game framework, the concept co-occurrence frequency metric, and the \dataset dataset.
This hidden language reveals the bias from imbalanced training, tracks system generalization progress, and helps improve concept presentation in systems' output.
We also use Reasoning-LLMs to infer how the multimodal systems' hidden language understands and simulates the world.
Crucially, this is a test-time scalable framework: As the computation scales, an increasingly comprehensive multimodal hidden language world map will unfold in front of our eyes.

Due to large-scale systems' diverse output, our framework may be influenced by inherent randomness.
In our future works, we will continuously conduct our \game to complete the hidden language world map, alleviating the impact of randomness.
We will also advance on directed path formation to support tasks in specific domains.
Additionally, we will apply various graph algorithms to this graph-based world map to identify optimal pathways between concepts.

\newpage
\appendix


\section{Metric Correlation} \label{sec:appendix/correlation}
\begin{figure}[htbp]
    \centering
    \subfigure[OpenAI System]{
        \includegraphics[width=0.31\textwidth]{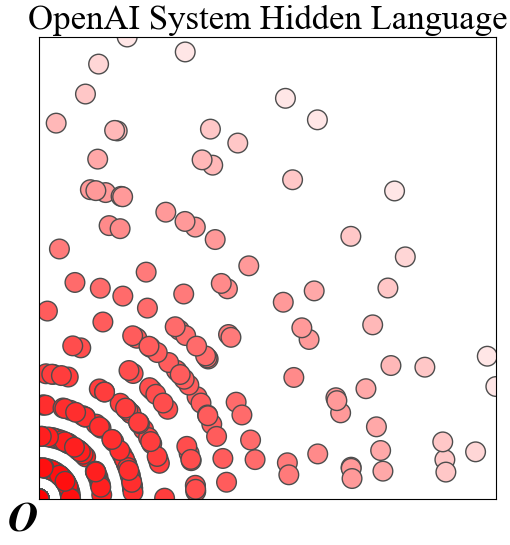}
        \label{fig:openai_hidden_language}
    }
    \subfigure[Semantic Similarity]{
        \includegraphics[width=0.31\textwidth]{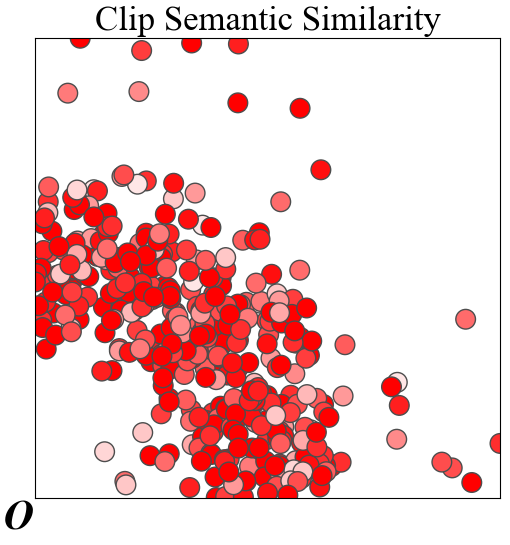}
        \label{fig:clip_sim}
    }
    \subfigure[Visual Similarity]{
        \includegraphics[width=0.31\textwidth]{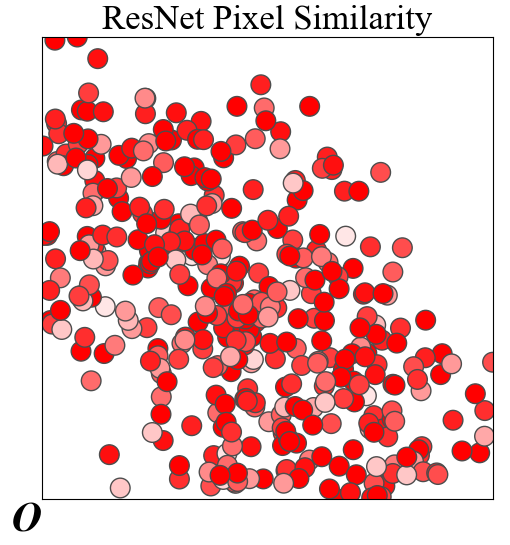}
        \label{fig:visual_sim}
    }
    \caption{The correlation between the 3 metrics. Each point represents a concept pair. The distance from the origin, $O$, indicates their similarity under the current metric, and its color intensity reflects the strength of their connection in the hidden space of the multimodal system, i.e., OpenAI System.}
    \label{fig:metric_correlation}
\end{figure}

In our Main Paper Section~\ref{sec:exp/correlation}, we introduce the quantitative correlation between the co-occurrence frequency, semantic and visual similarity.
Here, we present an intuitive scatter plot where each point represents the connection between two concepts.
The distance from the origin point $O$ indicates their similarity under a given metric, while color intensity reflects their connection strength in the multimodal hidden space (i.e., co-occurrence frequency).
For better visualization, concept pairs with the same similarity are randomly placed along the same-radius arc around the origin point $O$, rather than overlapping at a single point—resulting in a clearer 2D scatter plot.
We observe that neither semantic nor visual similarity can adequately explain the concept connections in multimodal systems.
\section{Visualization} \label{sec:appendix/visual}
\begin{figure}[htbp]
    \centering
    \includegraphics[width=\textwidth]{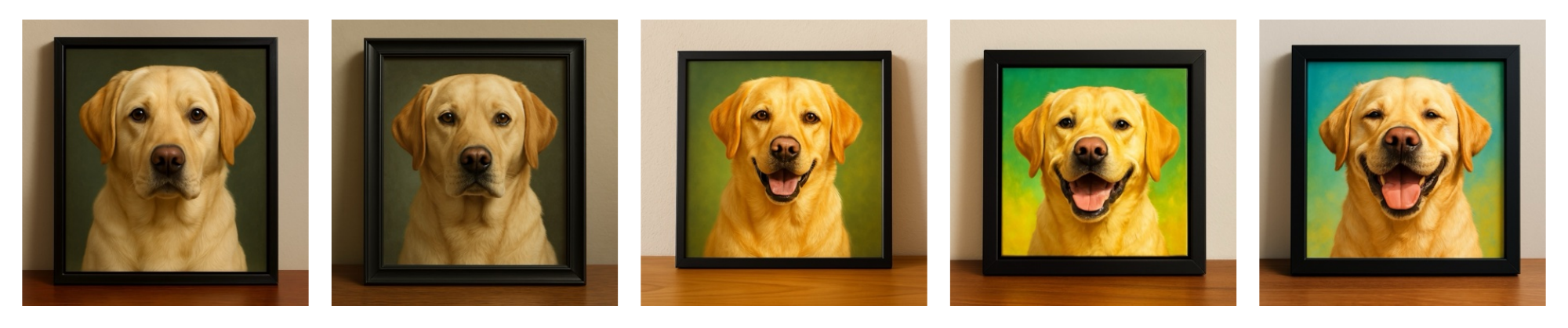}
    \caption{The extended \game of the teaser example in the Main Paper Figure~\ref{fig:teaser}, and the results eventually stabilize with the "dog in a frame."}
    \label{fig:appendix/extend_teaser}
\end{figure}
In this section, we will present the visualization results during the \game.
First, in Figure~\ref{fig:appendix/extend_teaser}, we extend the teaser example of \newgpt shown in our Main Paper Figure~\ref{fig:teaser} by continuing the \game.
The results eventually stabilize with the \say{dog in a frame,} indicating that the connection between \say{frame} and \say{dog} is indeed stronger than that between \say{TV} and \say{dog}.

\begin{figure}[htbp]
    \centering
    \includegraphics[width=\textwidth]{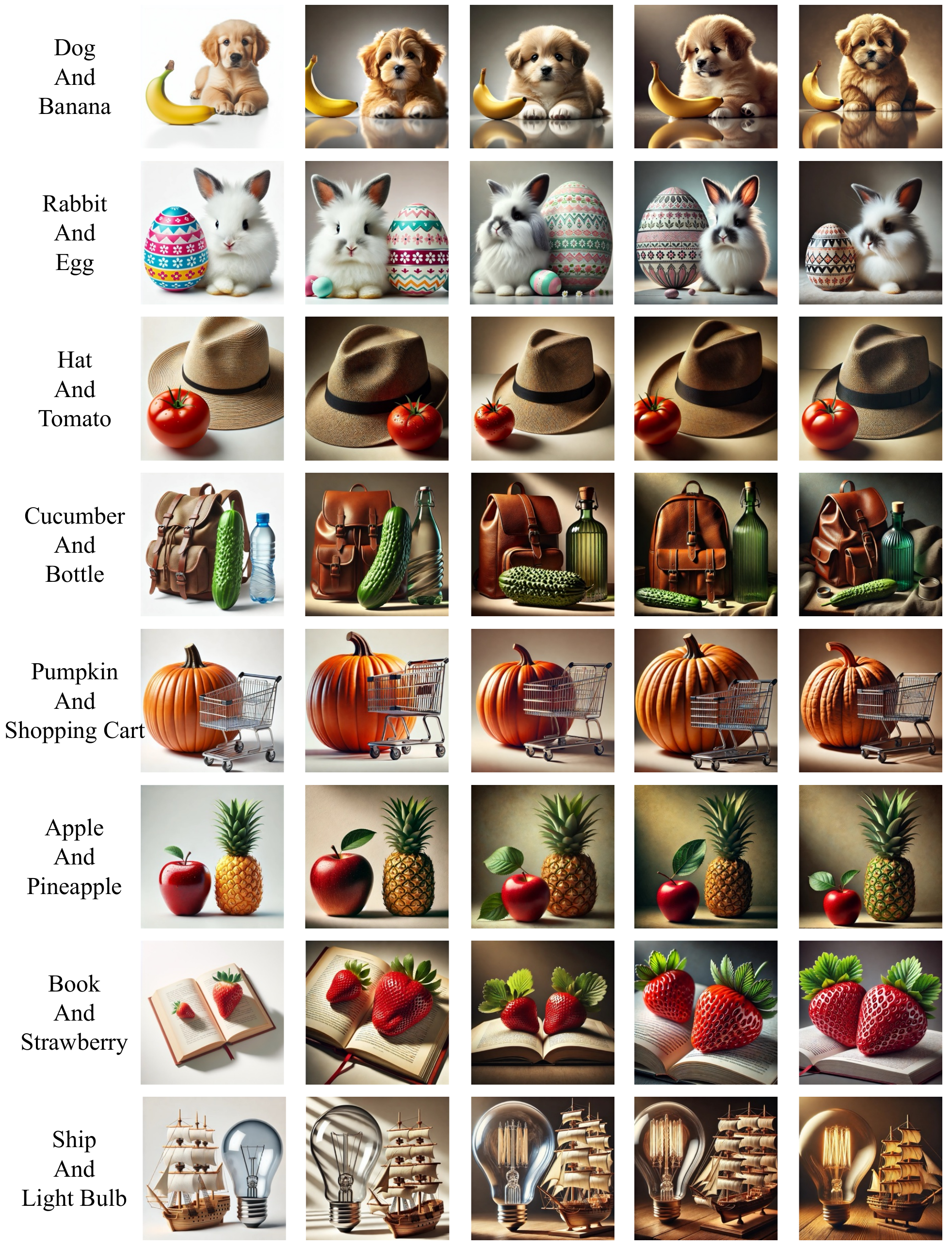}
    \caption{Several visualization results of concept pairs that remain stable during \game in our experiments, indicating strong concept connections in the hidden space.}
    \label{fig:appendix/good}
\end{figure}

\begin{figure}[htbp]
    \centering
    \includegraphics[width=\textwidth]{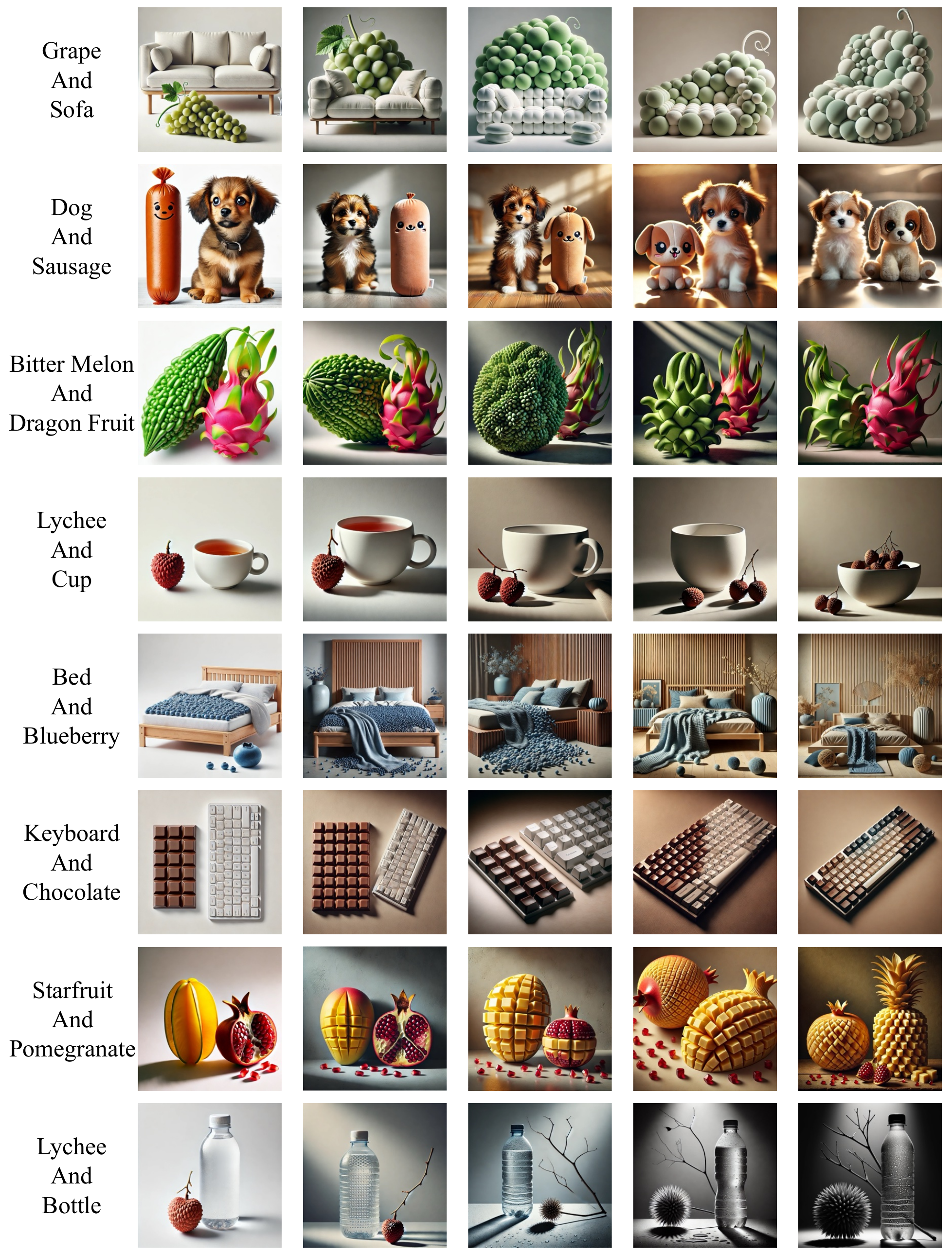}
    \caption{Several visualization results of concept pairs that exhibit concept crashes during \game in our experiments, indicating fragile concept connections in the hidden space.}
    \label{fig:appendix/bad}
\end{figure}

In Figure~\ref{fig:appendix/good}, we present additional concept pairs that remain stable during \game in our experiments, indicating strong concept connections in the multimodal hidden space, and Figure~\ref{fig:appendix/bad} shows more examples of concept pairs that exhibit concept crashes, indicating fragile concept connections in the multimodal hidden space.

\newpage

\section{LLMs as "MLPs"}
\label{sec:llm_as_mlp}
\begin{figure}[t]
    \centering
    \includegraphics[width=\textwidth]{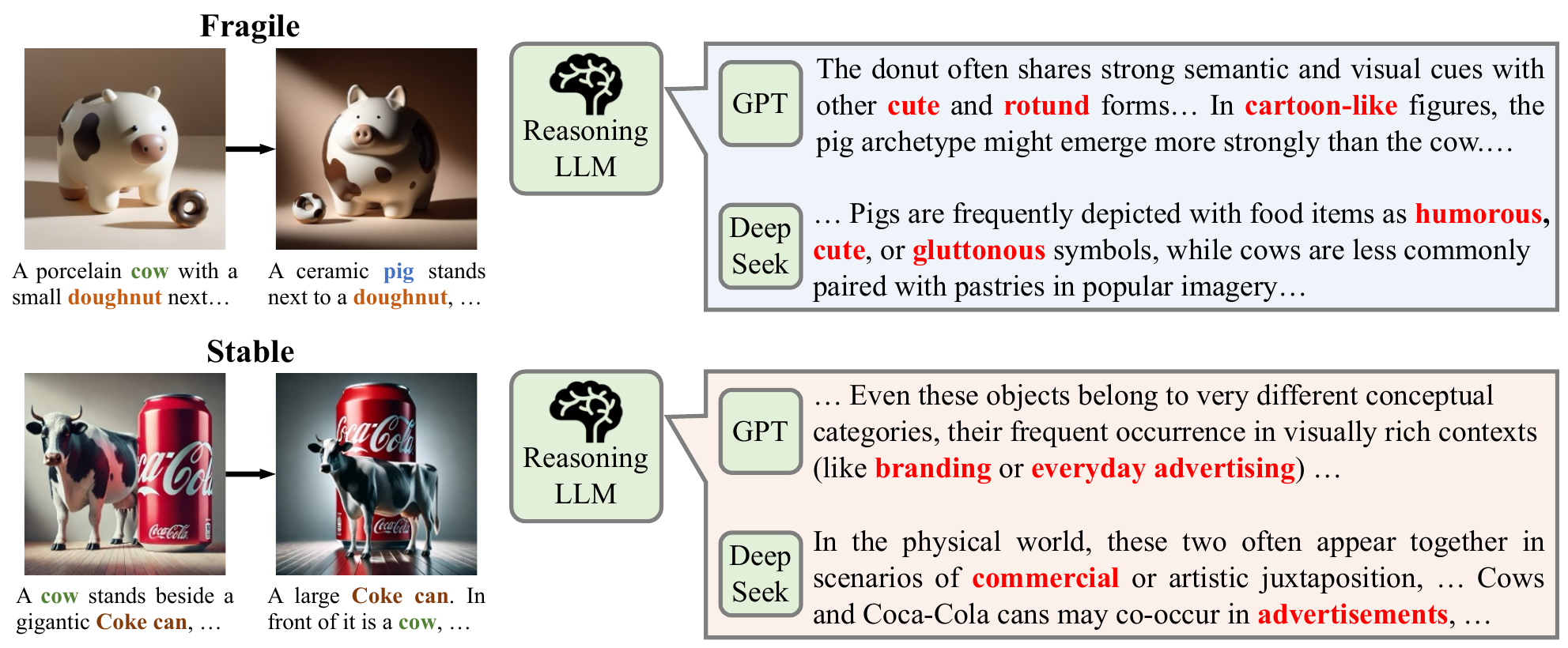}
    \caption{For hidden language beyond semantic or visual explanation, the LLMs Reasoning Analysis offers valuable insights into how multimodal systems understand the world.}
    \label{fig:reasoning}
\end{figure}

We employ the Reasoning-LLMs and try to explore the system’s hidden language about how it understands and simulates real-world physical laws.
We use 2 SOTA Reasoning-LLMs: GPT-o1~\citep{jaech2024openai} and DeepSeek-R1~\citep{guo2025deepseek}.
In Figure~\ref{fig:reasoning}, we show 2 examples representing fragile and stable connections.
Leveraging their reasoning capabilities, Reasoning-LLMs offer insights into the system's understanding of world patterns.
For example, milk and coke often co-occur in beverage areas, and cows appear frequently on milk packaging, leading to a stable connection between cow and coke.
These phenomena transcend semantic and visual similarities, revealing that multimodal systems are attempting to understand and simulate the common-sense laws in the human world.
For the instruction prompt, please refer to Appendix~\ref{sec:appendix/prompt}.

\begin{figure}[htbp]
    \centering
    \includegraphics[width=\textwidth]{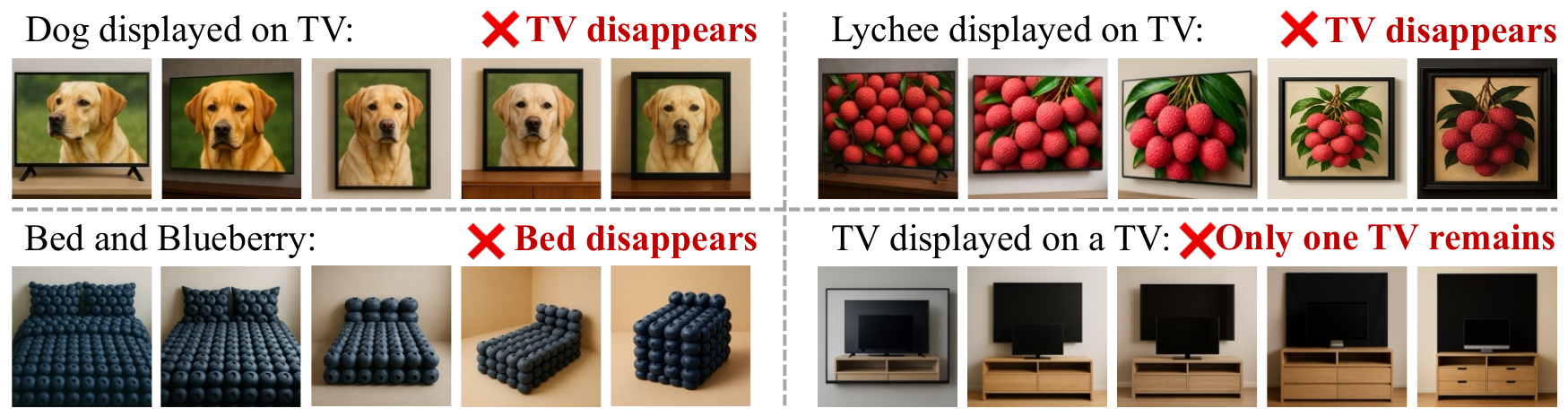}
    \caption{Visualization examples using the latest \newgpt. Despite its enhanced prompt-following capabilities, the system still reveals certain preference biases.}
    \label{fig:gen4o}
\end{figure}

\section{The Latest Released System} \label{sec:exp/gen4o}
OpenAI has released the \newgpt, claiming improved prompt-following ability and deep integration with \gpt.
This system is highly representative, transforming \gpt into a completely multimodal black-box system.
In Figure~\ref{fig:gen4o}, we conduct experiments on some representative examples using the web version.
We observe that it still exhibits certain preference bias during the \game, which will aid our exploration of its hidden language.

\newpage

\section{Connection Graph Structure} \label{sec:appendix/map}
As discussed in our Main Paper Section~\ref{sec:exp/connection}, we can construct an increasingly comprehensive \say{hidden language world map} of the multimodal system’s hidden space by accumulating more and more telephone games.

In this graph structure, each node represents a concept, and the edges between nodes indicate the connection strength between concept pairs in the hidden space.
These strengths are quantified using co-occurrence frequencies derived from a large number of repeated telephone games.
Specifically, the co-occurrence frequency between two concepts, $A$ and $B$, is calculated only from telephone games where $A$ and $B$ are present in the initial input, or newly emerged together during the process.

We can apply various graph algorithms~\citep{dijkstra2022note, floyd1962algorithm, bellman1958routing} to this \say{world map} structured as a graph to find optimized paths between concepts, enhancing their connections to be more natural and stable in the multimodal system's output.
For example, in our Main Paper Section~\ref{sec:exp/complex_pattern}, we observe that many concepts fail to appear consistently on a TV.
By examining the local graph map centered around the concept \say{TV}, we find that using the concept \say{cartoon style} as an intermediary node significantly improves the average connection strength.
Depending on the number of additional elements we want to introduce, we can select multiple intermediary nodes and then apply graph algorithms such as the shortest path (the stronger the connection, the shorter the distance between 2 concepts) to identify strong bridges between concepts.

\section{Open-Sourced Systems} \label{sec:appendix/opensource}
We use features from open-source models to observe and validate that, during the process of compressing images into text and then reconstructing them into images, the system tends to favor concept combinations with stronger connections in its hidden language.
We utilize the CLIP model~\citep{radford2021learning}, one of the fundamental components in open-source multimodal systems, to extract features from both images and text, which are then used to compute the similarities.

\begin{figure}[htbp]
    \centering
    \includegraphics[width=\textwidth]{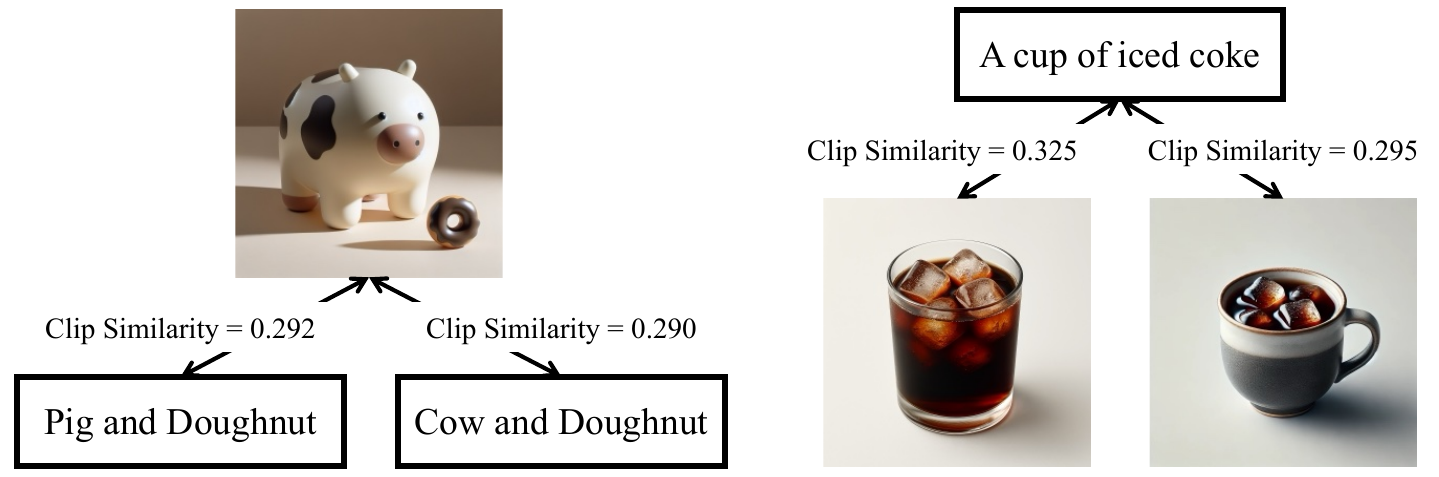}
    \caption{Preference bias of multimodal systems towards input concept combinations.}
    \label{fig:appendix/opensource}
\end{figure}

During the image-to-text compression process, concepts in the input image shift toward stronger concept combinations in the hidden space.
For example, in Figure~\ref{fig:appendix/opensource}, the systems prefer to interpret the image of \say{cow and doughnut} as a feature more similar to \say{pig and doughnut,} probably because pigs are more closely associated with doughnuts during training~\citep{zhao2024lost}.

During the text-to-image reconstruction process, concepts in input text shift toward stronger concept combinations in the hidden space.
For example, in Figure~\ref{fig:appendix/opensource}, the systems prefer to interpret the text of \say{a cup of iced coke} as a feature more similar to the scene of a glass of iced coke, probably because glasses are more closely associated with iced coke during training~\citep{zhao2024lost}.

\newpage

\section{Instruction Prompt} \label{sec:appendix/prompt}
Here we present the instruction prompt used in \game and Reasoning-LLMs.

\subsection{Get Start Description}
As discussed in Section~\ref{sec:game}, LLMs convert concept pairs in our \dataset dataset into the start description for our \game.
Here is the instruction prompt (To avoid interference from quotation marks in the system's response, we did not require it to output in JSON format.):
\begin{tcolorbox}[colback=blue!5!white,colframe=blue!75!black]
\# Task

Execute the text-to-image prompt generation task under these protocols:

\ 

\#\# Context Reset

- Two concepts the user provide are \{A\} and \{B\}

- The user wants to get a prompt describing these two concepts for creating an image

\ 

\#\# Specifications  

- If \{A\} and \{B\} are too different in SIZE, describe the scene as \{A\} and \{B\} are positioned next to each other. The background should be pure white without any other content.

- Else, describe the scene where \{A\} and \{B\} are most suitable when appearing. The background should be pure white without any other content.

- Refuse any detail or atmosphere description

\ 

\#\# Compliance Requirements  

- Strictly adhere to AI ethics guidelines

- Auto-filter NSFW/prohibited content

- Maintain objective observational stance

\ 

\# Response Format 

Only return your generated description, do not say anything else!
\end{tcolorbox}

\newpage

\subsection{Get Image Description}
In \game, we need the multimodal systems to compress the images into text descriptions.
Here is the instruction prompt (To avoid interference from quotation marks in the system's response, we did not require it to output in JSON format.):
\begin{tcolorbox}[colback=blue!5!white,colframe=blue!75!black]
\# Task

Execute the visual description task under these protocols:

\ 

\#\# Description Specifications
- Generate single-paragraph visual depiction

- Apply 'object + attributes + spatial relationships' framework

- Include comprehensive detailed visual elements:

\ \ \ \ -- All of the details in the image

\ \ \ \ -- Chromatic composition \& material textures

\ \ \ \ -- Lighting dynamics \& spatial perspective

\ \ \ \ -- Stylistic signatures \& atmosphere creation

\ 

\#\# Compliance Requirements

- Strictly adhere to AI ethics guidelines

- Auto-filter NSFW/prohibited content

- Maintain objective observational stance

\ 

\# Response Format

- Begin directly with the description, no phrases like 'The image shows or describes or presents'

- Only return your description, do not say anything else!
\end{tcolorbox}

\newpage

\subsection{Check Concept Existence}
In the co-occurrence frequency metric, we use LLMs to check if a concept exists in the image description. Here is the instruction prompt:

\begin{tcolorbox}[colback=blue!5!white,colframe=blue!75!black]
\# Task

Execute the Concept Existence Identify task under these protocols:

\ 

\#\# Context Reset

- Focus exclusively on the currently provided description

- You should check the existence of the concept, and check if new concepts appear in the description, instead of the background

- Provided description: '\{description\}'

- Concepts I want to check: '\{concept list\}'

\ 

\#\# Note

- You should first check if the concepts I want to check :'\{concept list\}' appear in the description scene, using True or False to indicate (If a concept I want to check is described by an alias, we consider it to appear)

- If there are new OBVIOUS concepts (not include background), add a new concept in return and use True to indicate

- Before adding any new concept, you must check it first:

\ \ \ \ -- SUBspecies: Any concept that are of the same species or subspecies as existing concepts: '\{concept list\}' should NOT be considered as a new concept. For example, cat and dog are different, but cat and ragdoll cat are the same. Cow and horse are different, but cow and bull are the same. Turtle and tortoise are the same.

\ \ \ \ -- Breed: If the new concept is a breed name of an existing concept: '\{concept list\}', it should NOT be considered as a new concept. For example, golden retriever is not a new concept (because it is the same as dog).

\ \ \ \ -- Different Term: Different words used to represent different ages of existing concepts: '\{concept list\}', it should NOT be considered as a new concept. For example, kitten is not a new concept (because it is the same as cat), and puppy is not a new concept (because it is the same as dog).

\ \ \ \ -- Background: The background and environment should NOT be considered as a new concept.

\ \ \ \ -- Light: The light of should NOT be considered as a new concept.

\ \ \ \ -- Part: The part of an existing concept's body, should NOT be considered as a new concept.

\ \ \ \ -- Style: The style/sense/feeling of the whole image should NOT be considered as a new concept.

\ \ \ \ -- Texture: The texture of an existing concept, should NOT be considered as a new concept.

\ \ \ \ -- Representation: The representation of an existing concept, such as a painting, a sculpture, a toy,etc., should NOT be considered as a new concept.

\ 

\# Response Format

- Only return a json, do not say anything else!

- The json format:

'''

\{

\ \ \ \ "\{ori\_concept\}": your decision (True or Fasle)

\}

'''

- If you find a new concept, you must check if it is a new concept. And if it is a new concept, add it to the return json.
\end{tcolorbox}

\newpage

\subsection{Reasoning LLMs}
We use Reasoning-LLMs to get insights into how multimodal systems understand the world.
Taking the example of \say{Cow and Coke}, here is the instruction prompt:

\begin{tcolorbox}[colback=blue!5!white,colframe=blue!75!black]
\# Task

Execute the Concepts Potential Connections Reasoning task under these protocols:

\ 

\#\# Background

- I am experimenting with a multimodal system that uses the "Pre-Description" to generate an image and then uses text to describe the image to get the "New Description".

- We find that in the above image reconstruction process, due to the system's existing preferences, it tends to reconstruct the concept combination that is more closely related in its understanding, which we call the Hidden Language of multimodal systems.

- The "Pre-Description": A majestic black and white Holstein cow stands confidently beside a gigantic red Coca-Cola can, casting distinct shadows on the hardwood floor below. The cow's coat boasts a clean sheen, highlighting its stark contrasts between patches of black and white, while its expressive eyes and symmetrical horns add to its regal presence. The Coca-Cola can towers over the cow, flaunting a vibrant red hue that dominates the scene with its glossy metallic texture reflecting bright lighting. The iconic white script logo of Coca-Cola is embossed on the can's surface, creating a strong branding visual impact. The scene features dynamic lighting that illuminates details on both subjects, generating subtle reflections and emphasizing seamless spatial coexistence in the serene, minimalist atmosphere.

- The "New Description": A large Coca-Cola can, predominantly bright red with the signature white logo curving across its shiny metallic surface, stands upright as an imposing backdrop. In front of this towering can is a black and white cow, which appears proportionally smaller, painted with a realistic sheen on its smooth, glossy skin, suggesting a polished, almost plasticine texture. The cow's body casts a long, soft shadow against the reflective hardwood floor, suggesting a light source positioned above and slightly in front of the duo, which faintly illuminates the details such as the cow's textured coat and the can's metallic glint. The setting exudes a surreal, hyperrealistic atmosphere, achieved through the juxtaposition of everyday elements at contrasting scales, creating a unique blend of reality and artistic abstraction.

\ 

\#\# Experimental Findings

- I find that two seemingly unrelated concepts, "Cow" and "Coke Can", are retained stably.

\ 

\#\# My Needs

- Please infer and analyze the reasons why these two seemingly unrelated concepts are closely related from the perspective of the multimodal system hidden space. What laws in the physical world it may reflect.

\ 

\# Response Format

'''

\{

\ \ \ \ "Reason": Describe your reasons.

\}

'''
\end{tcolorbox}

\newpage

\end{document}